\documentclass[runningheads]{llncs}
\usepackage{graphicx}
\usepackage{subcaption}
\usepackage[sectionbib]{natbib}
\usepackage{comment}
\begin{document}
\title{Discovering the Rationale of Decisions:\\ Experiments on Aligning Learning and Reasoning}

\titlerunning{Discovering the Rationale of Decisions}
%
\author{Cor Steging\inst{1} \and
Silja Renooij\inst{2} \and \\
Bart Verheij\inst{1}}
\authorrunning{Steging et al.}
%
\institute{Bernoulli Institute of Mathematics, Computer Science and Artificial Intelligence, University of Groningen \and
Department of Information and Computing Sciences, Utrecht University}
\maketitle              
\begin{abstract}
In AI and law, systems that are designed for decision support should be explainable when pursuing justice. In order for these systems to be fair and responsible, they should make correct decisions and make them using a sound and transparent rationale. In this paper, we introduce a knowledge-driven method for model-agnostic rationale evaluation using dedicated test cases, similar to unit-testing in professional software development. We apply this new method in a set of machine learning experiments aimed at extracting known knowledge structures from artificial datasets from fictional and non-fictional legal settings. We show that our method allows us to analyze the rationale of black box machine learning systems by assessing which rationale elements are learned or not. Furthermore, we show that the rationale can be adjusted using tailor-made training data based on the results of the rationale evaluation. 

\end{abstract}

\section{Introduction}
In AI and Law, explainability is a key requirement in system design, due to the need for the justification of decisions. 
For machine-supported decisions, this is nowadays encoded in the GDPR's right to explanation. Four types of explanations are distinguished~\citep{MILLER20191} and have been applied in AI and Law~\citep{atkinson2020explanation}: 
Contrastive explanations show why a decision is made and others are not. Examples include HYPO's counterexamples and hypothetical situations~\citep{HYPO,ashley1990} and argument diagrams~\citep{verheij2003aaa}. In selective explanations, the focus is on the most salient elements needed, for instance by the use of the critical questions of argumentation schemes~\citep{atkinson2020memoriam,verheij2003dialectical}. Probabilistic explanations are grounded in statistical correlations, and are less applicable in law with its focus on specific circumstances. An example is the explanation of evidential Bayesian networks~\citep{vlek2016method} in terms of scenarios and the evidence for and against them. Lastly, social explanations emphasise the transfer of knowledge between individuals, as in models of the dialogue between parties and in courts, specifying shared and unshared commitments~\cite{hageLeenesLodder1993,gordon1995,atkinson2020memoriam}. 

This requirement of explainability is problematic for the application of central machine learning techniques in law. Neural networks, for example, are known to perform well, but behave like a black box algorithm. Hence, explanation techniques have been developed to `open the black box' (cf. LIME \citep{lime}, SHAP \citep{NIPS2017_7062}). Even in the domain of vision (where the successes of neural networks are especially significant), the necessity of such methods is underpinned by studies regarding adversarial attacks that show that slight perturbations of images, invisible to the human observer, can radically change the outcome of a classifier~\citep{goodfellow2014explaining}. 

In this paper, we expand upon the method introduced in \citep{ICAILpaper}, where we investigate black box machine learning methods with a focus on proper explainability, and not only in terms of accuracy as in the standard machine learning protocol. We are in particular interested in the discovery of the rationale underlying decisions, where the rationale is the knowledge structure that can justify a decision, such as the rule applied. We aim to measure the quality of rationale discovery, with an eye on the possibility of improving rationale discovery. 

To measure and possibly improve rationale discovery, we create dedicated test datasets, on which a machine learning system can only perform well if it has learned a particular component of the knowledge structure that defined the data. The idea is similar to how unit testing works in professional software development: we define a set of cases, targeting a specific component, in which we know what the answer should be, and compare that to the output that the system gives. 

To be able to focus on what is methodologically feasible, we do not use natural language corpora (as for instance in argument mining~\citep{mochalesMoens2009,wynerEtal2010}, conceptual retrieval~\citep{grabmairEtal2015} or case prediction~\citep{ashley2019,medvedevaEtal2019,bruninghausAshley2003}). Instead we work with datasets of artificial decisions with known underlying generating rationale. 

Our work builds on a study investigating whether neural networks are able to tackle open texture problems~\citep{bench1993neural}. The study used a fictional legal domain (also investigated in \citep{wardeh2009padua, movzina2005argument}), in which the eligibility for a welfare benefit for elderly citizens is determined based on six conditions. Artificial datasets were generated specifying personal information of elderly citizens with their eligibility for the welfare benefit. Multilayer perceptrons were trained and tested on these datasets, and managed to perform with high accuracy scores (above 98\%). It was shown that the neural networks were unable to properly learn two of the six conditions. By making adjustments to the training dataset, the neural networks were able to learn conditions more adequately, while maintaining similar accuracy scores. But also after adjustment, the conditions that defined the data were not learned fully correctly. Other earlier discussions of neural networks in law are~\citep{philipps1999introduction,hunter1999out,stranieriEtal1999}. 

In the upcoming sections, the study by \citet{bench1993neural} will first be replicated as closely as possible, using modern, widely-used neural network methods, in order to affirm whether the claims made in 1993 still hold today. Then a simplified version of the welfare benefit domain is examined to see how well the networks are able to extract a simplified rationale. Lastly, we study a real legal setting, namely Dutch tort law. That domain uses only Boolean variables, but allows for exceptions to underlying rules.

\section{Domains and Datasets}
For each of the three domains considered in this paper, this section describes the underlying knowledge structure using logic, from which we will generate datasets to train a series of neural networks. These networks will subsequently be analysed using a method we propose for assessing the quality of their rational discovery. To this end we need two types of datasets for the purpose of testing. The first are standard test sets sampled from the complete domain to evaluate the accuracy of the networks. The second type is a dedicated test set designed to target a specific aspect of the domain knowledge. This section describes all datasets we use. 

\subsection{Domains}
\subsubsection{Welfare benefit domain}
\noindent  This fictional domain introduced in \citep{bench1993neural} concerns the eligibility of a person for a welfare benefit to cover the expenses for visiting their spouse in the hospital, and can be formalised as follows: \\[-5pt]

\noindent \begin{tabular}{l@{}l@{}l}
$Eligible(x)$ &  $\iff$ & $C_1(x) \land C_2(x) \land C_3(x) \land C_4(x) \land C_5(x) \land C_6(x)$\\
\end{tabular}

\noindent\begin{tabular}{l@{}l@{}l}
$C_1(x)$ &  $\iff$ & $(Gender(x) = female \land Age(x) \geq 60) \lor$ \\
         &         & $(Gender(x) = male \land Age(x) \geq 65)$ \\
$C_2(x)$ &  $\iff$ & $|Con_1(x), Con_2(x), Con_3(x), Con_4(x), Con_5(x) | \geq 4 $\\
$C_3(x)$ &  $\iff$ & $Spouse(x)$ \\
$C_4(x)$ &  $\iff$ & $\neg Absent(x)$ \\
$C_5(x)$ &  $\iff$ & $\neg Resources(x) \geq 3000$ \\
$C_6(x)$ &  $\iff$ & $(Type(x) = in \land Distance(x) < 50) \lor$ \\
         &         & $(Type(x) = out \land Distance(x) \geq 50)$\\[5pt]
\end{tabular}

\noindent
That is, a person is eligible iff he/she is of pensionable age (60 for a woman, 65 for a man), paid four out of the last five contributions $Con_i$, is the patient's spouse, is not absent from the UK, has capital resources not amounting to more than \pounds3,000, and lives at a distance of less than 50 miles from the hospital if the relative is an $in$-patient, or beyond that for an $out$-patient.

\begin{table}[t]
\centering
\caption{Features in the welfare benefit domain.}
\begin{tabular}{ll}
\textbf{Feature}            & \textbf{Values} \\ \hline
$Age$                       & 0 -- 100 (all integers)               \\
$Gender$                    & male or female       \\
$Con_1$,\ldots, $Con_5$     & true or false        \\
$Spouse$                    & true or false        \\
$Absent$                    & true or false        \\
$Resources$                 & 0 -- 10,000 (all integers)        \\
$Type$ (Patient type)               & in or out            \\
$Distance$ (to the hospital)        & 0 -- 100 (all integers)              
\end{tabular}
\label{tbl:features}
\end{table}

The six \emph{independent} conditions for eligibility are defined in terms of 12 variables, which are the features of the generated datasets. These features and their possible values are shown in Table \ref{tbl:features}. 
In addition to these 12 features, the datasets will contain 52 noise features unrelated to eligibility, just as in the original experiment, giving a total of 64 features plus an eligibility label for each instance. All datasets are valid in the sense that the given eligibility labels follow from evaluating the 6 conditions above.

\subsubsection{Simplified domain}
Experiments with different models for the welfare domain all concluded that it was not possible to extract all six conditions for eligibility \citep{bench1993neural,wardeh2009padua,movzina2005argument}. The complexity of the original problem with 6 different conditions and 64 features, complicate a proper analysis of the networks' rationale, since each condition and feature could potentially influence it. To facilitate this analysis, we simplified the original problem in two ways. First, the 52 noise variables, which did not seem to affect the performance of the networks \citep{bench1993neural}, are removed. Secondly, we define eligibility solely by the age-gender ($C_1$) and patient-distance ($C_6$) conditions that were examined in the original experiment to justify the rationale of the network: \[Eligible(x) \iff C_1(x) \land C_6(x)\]
Eligibility is thus determined through a combination of a XOR-like function ($C_6$) and a nuanced threshold function ($C_1$).

\subsubsection{Tort law domain}
Our third domain concerns Dutch tort law: articles 6:162 and 6:163 of the Dutch civil code that describe when a wrongful act is committed and resulting damages must be repaired. This `duty to repair' ($dut$) can be formalised as follows: \\[-5pt]

\begin{figure}[tb]
    \centering
    		 \includegraphics[width = 0.9\linewidth]{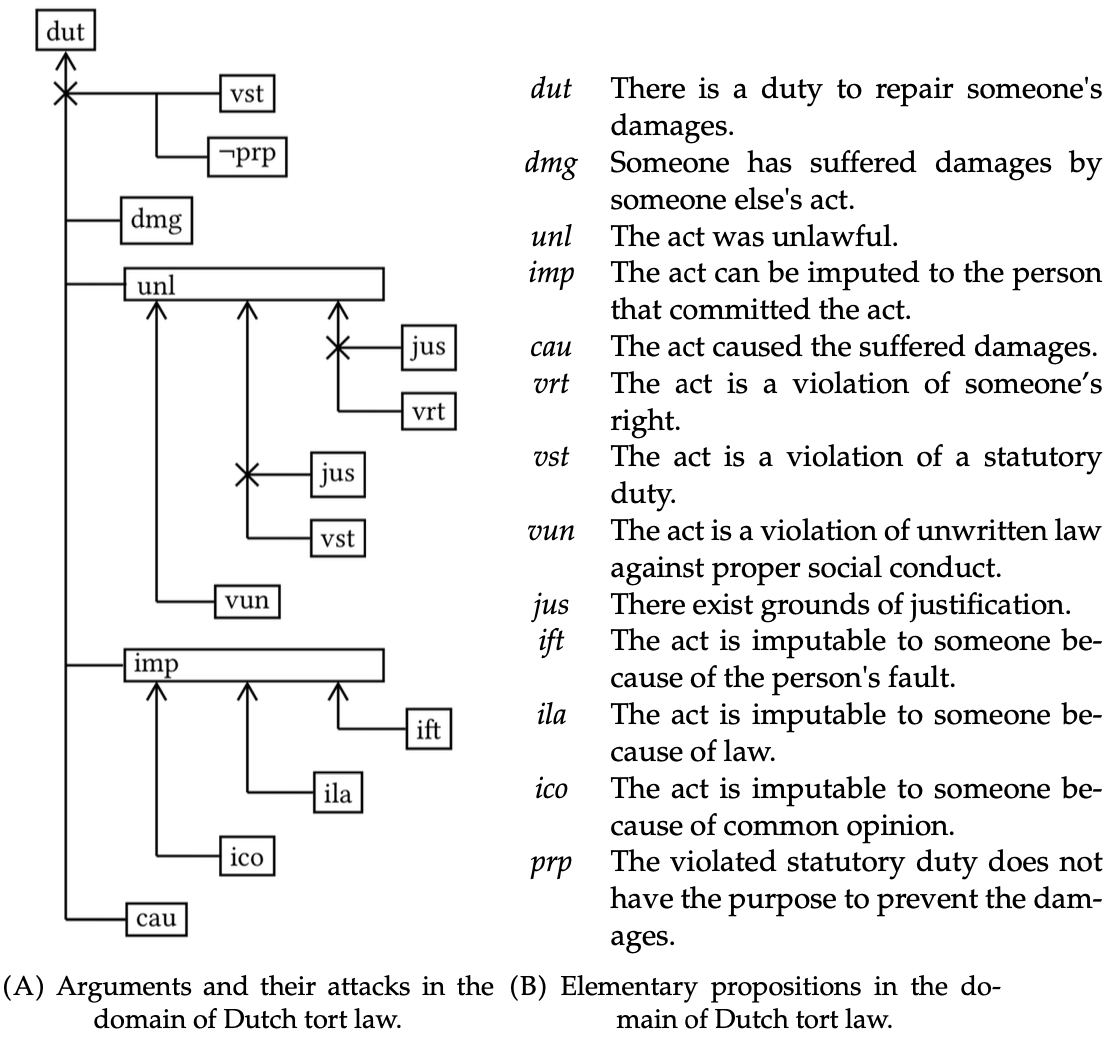}
    \caption{Arguments and attacks (A) and their elementary propositions (B) in 
    Dutch tort law \citep{verheij2017formalizing}.}
    \label{fig:TortDiagram}
\end{figure}

\noindent\begin{tabular}{l@{}l@{}l}
$dut(x)$ & $\iff$ & $c_1(x) \land c_2(x) \land c_3(x) \land c_4(x) \land c_5(x)$\\
$c_1(x)$ & $\iff$ &  \textit{cau}$(x)$\\
$c_2(x)$ & $\iff$ & \textit{ico}$(x)\ \lor$ \textit{ila}$(x)\ \lor$ \textit{ift}$(x)$\\
$c_3(x)$ & $\iff$ & \textit{vun}$(x)\, \lor ($\textit{vst}$(x)\, \land \neg$\textit{jus}$(x))\, \lor ($\textit{vrt}$(x)\, \land \neg$\textit{jus}$(x))$\\
$c_4(x)$ & $\iff$ &  \textit{dmg}$(x)$\\
$c_5(x)$ & $\iff$ & $\neg ($\textit{vst}$(x)\ \land \neg$\textit{prp}$(x))$\\[5pt]
\end{tabular}

\noindent
where the elementary propositions are provided alongside an argumentative model  of the law in Figure~\ref{fig:TortDiagram}  \citep{verheij2017formalizing}, and conditions $c_2$ and $c_3$ capture the legal notions of unlawfulness (\textit{unl}) and imputability (\textit{imp}), respectively. 

Compared to the fictional welfare domain, the Dutch tort law domain is captured in 5 conditions for duty to repair (\textit{dut}), based upon 10 Boolean features. Each condition is a disjunction of one or more features, possibly with exceptions. The feature capturing a violation of a statutory duty (\textit{vst}) is present in both condition $c_3$ and $c_5$, rendering these \emph{dependent}. 

\subsection{Datasets}
For each experiment, we generate datasets of different types, for different purposes\footnote{The Jupyter notebooks used for data generation can be found in a Github repository: \url{https://github.com/CorSteging/DiscoveringTheRationaleOfDecisions}}
For most types of datasets, the generating process is at least partly stochastic and repeated for every repetition of an experiment. Using the same \textit{type} of dataset, for example in training and testing a neural network, does therefore not mean using the exact \textit{same} dataset. 
Table \ref{tbl:overview_domains} shows an overview of the domains and their datasets, and illustrates their differences and similarities.

\subsubsection{Welfare benefit datasets}
Within this domain, four types of datasets are generated, following the original study \citep{bench1993neural} as closely as possible: type A, type B, Age-Gender and Patient-Distance datasets. Each dataset contains the 12 features as defined in Table \ref{tbl:features}, as well as 52 noise features with integer values ranging from 0 to 100. The original study used training sets with 2,400 instances, which is quite small by todays standards \citep{atkinson2020explanation}. To make sure conclusions are not the result of using too little data, we will also include training sets with more data (50,000 instances). 

\emph{Type A datasets} are generated with either 2,400 instances or 50,000 instances. Exactly half of the instances are eligible, creating a balanced label distribution, as is common practice in machine learning problems. For the eligible instances, feature values are generated (randomly where possible) such that they satisfy the conditions $C_1-C_6$. For each condition, $\frac{1}{6}$th of the ineligible instances is designed to fail on that specific condition; where possible the values of the features involved are generated randomly such that the condition fails. All remaining features in these instances are generated randomly across their full range of values (see  Table \ref{tbl:features}); as a result, it is possible for ineligible instances to fail on multiple conditions, and some conditions will fail more often than others.  

In the original 1993 study it was argued that it was too easy to achieve high accuracy scores with networks trained and tested on type A datasets, which contained an average of 4.1 conditions that were not satisfied for ineligible cases. 
Using only 4 out of 6 conditions was shown to be sufficient for classifying 98.95\% of the instances correctly. 

\emph{Type B datasets} were subsequently introduced to make the problem more challenging. These datasets differ from type A datasets only in that ineligible instances fail on \emph{exactly} one condition, rather than at least one condition; the other five conditions are always satisfied. Type B datasets again contain either 2,400 instances or 50,000 instances. 

The original study investigated whether ``an acceptable rationale can be uncovered by an examination of the net'' \citep{bench1993neural}. To this end a set of test cases was constructed in which all conditions except one were guaranteed to be satisfied. From these it was concluded that the age-gender condition ($C_1$) and the patient-distance condition ($C_6$) are not learned by network. The original paper does not specify how many test cases were constructed, nor exactly how they were constructed. We will generate dedicated datasets that are tailor made to evaluate whether the trained networks have actually uncovered these same two conditions. 

The \emph{Age-Gender datasets} are generated by sampling the age and gender features across their full range of values, this time considering only multiples of 5 for age. The values for the other features are generated such that every condition is satisfied except for the age-gender condition ($C_1$). As a result, the eligibility of an instance in these datasets is solely determined by whether or not condition $C_1$ is satisfied. The Age-Gender sets contain 40,000 instances, with every possible combination of values for age and gender occurring a 1000 times. This gives a slightly unbalanced label distribution with 42.5\% of the instances being eligible, and 57.5\% ineligible. Because the dataset is only used to test the networks, rather than to train the network, this is not an issue. 

\begin{table}[!t]
\caption{An overview of the three domains and their datasets. Datasets marked with an asterisk are used for testing purposes only. For each type of dataset, the size and label distribution is given. For each domain is indicated the number and type of features, the number of type of conditions to be learned, whether or not all cases are covered by the datasets (complete) and whether the domain is fictional or real.}
\begin{tabular}{l|lccc}
  &         &   & \textbf{T/F label} & \textbf{Features:}\\[-1pt]
\textbf{Domain}          & \textbf{Dataset}          & \textbf{Size}  & \textbf{distribution}  &   \textbf{no. and type} \\
 \hline
Welfare benefit & Type A           & 2,400/50,000 & 50\%/50\%                   & 64       \\
                & Type B (fail on 1)         & 2,400/50,000 & 50\%/50\%         & Boolean \&               \\
                & Age-Gender*       & 40,000       & 42.5\%/57.5\%             & numerical               \\
                & Patient-Distance* & 40,000       & 50\%/50\%                 &                \\ \hline
Simplified      & Type A           & 50,000        & 50\%/50\%                 & 4               \\
welfare benefit & Type B (fail on 1)          & 50,000        & 50\%/50\%      & Boolean \&               \\
                & Age-Gender*       & 4,242       & 42.5\%/57.5\%              & numerical                          \\
                & Patient-Distance* & 3,234       & 50\%/50\%                  &                \\ \hline
Tort law        & Regular          & 5,000/500    & 50\%/50\%                  & 10              \\
                & Unique*           & 1024         & 10.94\%/89.06\%           & Boolean                \\
                & Unlawfulness*     & 168          & 66.67\%/33.33\%           &               \\
                & Imputability*     & 128          & 87.5\%/12.5\%             &               \\ \\ 
\textbf{Domain}          & \multicolumn{2}{l}{\textbf{Conditions: no. and type}}     & \textbf{Complete} & \textbf{Fictional}\\
 \hline
Welfare benefit & \multicolumn{2}{c}{6,\  Independent} & No             & Yes       \\
 \hline
Simplified      & \multicolumn{2}{c}{2,\ Independent} & No             & Yes       \\
welfare benefit &  \\
 \hline
Tort law        &\multicolumn{2}{c}{5,\ Dependent}  & Yes            & No        \\
\end{tabular}
\label{tbl:overview_domains}
\end{table}

The \emph{Patient-Distance datasets} are similarly generated by sampling the distance and patient type features across their full range of values, this time considering only multiples of 5 for distance. 
The eligibility of an instance in these datasets is thus determined by whether or not condition $C_6$ is satisfied. The Patient-Distance datasets also contain 40,000 instances, with every possible combination of values for patient type and distance occurring a 1000 times. In these datasets, exactly 50\% of the instances is eligible.

\subsubsection{Simplified datasets}
\noindent For the simplified welfare domain, the same type of datasets are generated as above, with the same properties except that all noise features and 8 of the 12 actual features are excluded. For type B datasets, this means that ineligible instances fail on either $C_1$ or $C_6$, but not on both, while in type A datasets the ineligible instances can fail on both conditions. Moreover, in the Age-Gender dataset the patient-distance condition $C_6$ is always satisfied, and in the Patient-Distance dataset the age-gender condition $C_1$ is always satisfied. Type A and type B datasets again contain 50,000 instances each. The Age-Gender dataset now contains only two features and 4,242 instances, that is, one unique instance for every possible combination of age and gender. Likewise, the Patient-Distance dataset contains 3,234 unique instances.

\subsubsection{Tort law datasets}
With 10 Boolean features there are $2^{10}=1024$ possible unique cases that can be generated from the argumentation structure of the tort law domain in Figure~\ref{fig:TortDiagram}. Each case has a corresponding outcome for \textit{dut}, indicating whether or not there is a duty to repair someone's damages. We will again consider four types of datasets. 

The \emph{unique dataset} contains these 1024 unique instances for the 10 features plus the label. In this dataset, there are 912 instances where \textit{dut} is false and 112 instances where \textit{dut} is true (11\%).

The \emph{regular type datasets} are generated such that \textit{dut} is true in exactly half of the instances. The sets are regular in the sense that balanced label distributions are common in machine learning problems. These regular datasets are generated by sampling uniformly from the subset of cases from the unique dataset, such that each possible case is represented equally within the 50/50 label distribution.
In practice, only a subset of the possible cases is typically available and presented to a network, upon which the network will have to learn to generalize to all possible cases. In addition to generating regular type datasets with 5,000 cases, we therefore also generate smaller regular type datasets with only 500 instances; the latter contains 35.35\% of the unique instances.

In the tort law domain we focus on the notions of unlawfulness ($c_2$) and imputability ($c_3$) to assess whether the networks are able to discover conditions in the data. For each of the two conditions, we again create a dedicated dataset. 

The \emph{Unlawfulness dataset} is the subset of the unique dataset in which the features for the unlawfulness condition $c_2$ can take on any of their values, while the other features have values that are guaranteed to satisfy the remaining conditions.Whether or not there is a duty to repair is therefore solely determined by whether or not condition $c_2$ is satisfied. All combinations of values of the other features are considered. The Unlawfulness dataset therefore consists of 168 unique instances, of which 66.66\% have a positive $dut$ value.

The \emph{Imputability dataset} is a similar subset of the unique dataset, but now the features for the imputability condition ($c_3$) can take on any value, provided that the value of \textit{vst} is such that condition $c_5$ is satisfied. The value of $dut(x)$ now completely depends on whether or not condition $c_3$ evaluates to true. Due to the interdependency of conditions $c_3$ and $c_5$, the Imputability dataset only has 128 unique instances, with 87.5\% of them having a positive $dut$ value. 

\section{Experimental setup and results}
In this section we describe and motivate the experiments we performed and report on their results.

\subsection{Experiments}
In order to demonstrate our method for assessing and improving rationale discovery of models learned from data, we first need such models. Though our method is model agnostic, we choose to use neural networks, like in \citep{bench1993neural}. 
\begin{table}[h!]
\centering
\caption{Experimental setup. For each domain, every listed training set is used in combination with all listed test sets.}
\begin{tabular}{@{\, }l@{\, }|@{\, }l@{\, }|@{\, }l@{}}
                & \textbf{Train on}         & \textbf{Test on}         \\ \hline
Welfare         & Type A (2,400 instances)  & Type A (2,400 instances) \\
benefit         & Type B (2,400 instances)  & Type B (2,400 instances) \\
                & Type A (50,000 instances) & Age-Gender               \\
                & Type B (50,000 instances) & Patient-Distance         \\ \hline
Simplified      & Type A                    & Type A                   \\
welfare         & Type B                    & Type B                   \\
benefit         &                           & Age-Gender               \\
                &                           & Patient-Distance         \\ \hline
Tort law        & Regular (5,000 instances)  & Regular (5,000 instances) \\
                & Regular (500 instances)   & Unique                   \\
                &                           & Unlawfulness             \\
                &                           & Imputability            
\end{tabular}
\label{tbl:overview_experiments}
\end{table}
We assume that assessing and improving rationale discovery is relevant only for models that are considered to be a good match with the data they were learned from. Our first step, after training the above mentioned neural networks, is therefore to evaluate their performance on typical test sets in terms of the standard accuracy measure. Subsequently we will evaluate the performance of the networks on the dedicated, knowledge-driven test sets that were specifically designed for assessing the networks' quality of rationale discovery.

\subsubsection{Neural network architectures}
In the original experiments in 1993, three multilayer perceptrons were used with one, two and three hidden layers, respectively \citep{bench1993neural}. These networks were created using the Aspirin software \citep{Leighton1994}, but the exact details regarding the networks and its parameters (e.g. the learning rate, activation function, gradient descent method) were left out of the original publication. The networks all had 64 input nodes (one for each feature in the datasets), a varying number of nodes in the hidden layers, and one output node that determines the eligibility. 

In this paper we will use a similar set-up and network architecture for all three domains. The output is always a single node, representing either eligibility or duty to repair, depending on the domain under consideration. The number of input nodes corresponds to the number of features and is therefore dependent on the domain (see Table~\ref{tbl:overview_domains}). More specifically, the welfare benefit domain will have 64 input nodes, the simplified domain will have 4, and the tort law domain will have 10 input nodes. The node configuration (i.e. number of nodes per layer) of each network is as follows, where \textit{input} represents the number of nodes in the input layer:

\begin{itemize}
    \item One hidden layer network: \textit{input}-$12$-$1$
    \item Two hidden layer network: \textit{input}-$24$-$6$-$1$
    \item Three hidden layer network: \textit{input}-$24$-$10$-$3$-$1$
\end{itemize}

\noindent In the replication of the 1993 experiment, the MLPClassifier of the scikit-learn package is used \citep{scikit-learn}. The networks use the sigmoid function as their activation function, which was the most common activation function when the original study was done. The networks use the Adam stochastic gradient-based optimizer \citep{kingma2014adam}, with a constant learning rate of 0.001. A total of 50,000 training iterations are used with a batch size of 50. Recall that the focus of this study is not on creating the best possible classifier, but to demonstrate our method of assessing rationale discovery.

\subsubsection{Training and performance testing}
The three types of neural networks will be trained and tested on a combination of different datasets, from each of the three domains. A complete overview of the datasets used in the experiments is shown in Table \ref{tbl:overview_experiments}. This table shows the datasets that the networks will train on, and the datasets that the networks will be tested with. For each domain, every \emph{combination} of training dataset and testing dataset is evaluated in terms of the accuracy of the resulting network on the test data. Because some of the datasets are stochastic (each generated dataset is slightly different), the whole process of data generation, training and testing is repeated 50 times. The mean classification accuracies along with their standard deviations will be reported.

To assess the rationale discovery capabilities of all the trained networks, we study their performance on the dedicated test sets for the age-gender, patient-distance, unlawfulness and imputability conditions. Performance will be measured both quantitatively, using standard accuracy, and qualitatively by a more detailed comparison of actual and expected outcomes.

\subsection{Results}
We will first report the accuracy scores for all combinations of training and testing datasets in the different domains and subsequently focus on rationale discovery. Results will be discussed in detail in Section~\ref{sec:discussion}.

\begin{table}[t!]
\centering
\caption{The accuracies obtained by the neural networks in the original study \citep{bench1993neural}.}
\begin{tabular}{l|cc|cc}
                         & \multicolumn{2}{c|}{\textbf{Trained on training set A}}   & \multicolumn{2}{c}{\textbf{Trained on training set B}} \\
                         & Test set A                & Test set B      & Test set A                & Test set B \\ \hline
1 hidden layer           & 99.25                     & 72.25           & 99.25                     & 97.91           \\ 
2 hidden layers          & 98.90                     & 76.67           & 99.00                     & 98.08            \\ 
3 hidden layers          & 98.75                     & 74.33           & unconverged               & unconverged          \\ 
\end{tabular}
\label{tbl:original_performance}
\end{table}
\begin{table}[t!]
\centering
\caption{The accuracies obtained by the neural networks in the replication study.}
\begin{tabular}{l|cccc}
                & \multicolumn{4}{c}{\textbf{Trained on training set A}}         \\
                & Test set A & Test set B & Age-Gender & Patient-Distance \\ \hline
1 hidden layer  & 98.97±0.19 & 72.39±1.66 & 52.14±4.01 & 50.05±0.09 \\
2 hidden layers & 98.87±0.21 & 72.56±1.83 & 53.19±4.56 & 50.06±0.14  \\ 
3 hidden layers & 98.92±0.23 & 70.97±1.74 & 50.45±3.23 & 50.03±0.07 
\end{tabular}
\begin{tabular}{l|cccc}
                & \multicolumn{4}{c}{\textbf{Trained on training set B}}         \\
                & Test set A & Test set B & Age-Gender & Patient-Distance \\ \hline
1 hidden layer  & 96.13±0.66 & 90.51±1.25 & 86.4±1.33 & 85.77±5.21 \\
2 hidden layers & 95.5±0.87  & 89.4±1.5  & 85.62±1.21 & 83.09±7.22 \\ 
3 hidden layers & 93.95±9.03 & 86.34±7.71 & 83.81±7.29 & 74.57±12.8
\end{tabular}
\label{tbl:new_performance}
\end{table}
\begin{table}[t!]
\centering
\caption{The accuracies obtained by the neural networks in the replication study with more training data.}
\begin{tabular}{l|llll}
                & \multicolumn{4}{c}{\textbf{Trained on training set A}}  \\
                & Test set A & Test set B & Age-Gender & Patient-Distance \\ \hline
1 hidden layer  & 99.8±0.03  & 80.98±1.47 & 60.22±3.87 & 64.44±2.87   \\
2 hidden layers & 99.79±0.04 & 83.49±1.86 & 65.04±5.12 & 66.5±3.24  \\
3 hidden layers & 99.78±0.05 & 82.89±2.05 & 64.45±5.6  & 64.2±3.1    
\end{tabular}
\begin{tabular}{l|llll}
                & \multicolumn{4}{c}{\textbf{Trained on training set B}}  \\
                & Test set A & Test set B & Age-Gender & Patient-Distance \\ \hline
1 hidden layer  & 99.64±0.17 & 98.53±0.15 & 98.51±0.47 & 97.17±0.46     \\
2 hidden layers & 99.28±0.36 & 98.06±0.35 & 97.75±0.9  & 96.53±0.3101     \\
3 hidden layers & 98.95±0.55 & 97.5±0.4   & 96.72±1.0  & 95.73±0.86    
\end{tabular}
\label{tbl:more_data_performance}
\end{table}
\begin{table}[t!]
\centering
\caption{The accuracies obtained by the neural networks in the simplified welfare domain.}
\begin{tabular}{l|llll}
                & \multicolumn{4}{c}{\textbf{Trained on training set A}}      \\
                & Test set A & Test set B & Age-Gender & Patient-Distance  \\ \hline
1 hidden layer  & 99.12±0.06 & 98.2±0.15  & 99.68±0.1 & 97.75±0.05      \\
2 hidden layers & 99.61±0.14 & 99.2±0.28  & 99.88±0.1 & 98.14±0.53 \\
3 hidden layers & 99.48±0.27 & 99.01±0.47 & 99.7±0.46 & 98.06±0.61 
\end{tabular}
\begin{tabular}{l|llll}
                & \multicolumn{4}{c}{\textbf{Trained on training set B}}      \\
                & Test set A & Test set B & Age-Gender & Patient-Distance  \\ \hline
1 hidden layer  & 99.46±0.06 & 99.04±0.12 & 99.67±0.06 & 98.04±0.31     \\
2 hidden layers & 99.77±0.14 & 99.6±0.25  & 99.78±0.18 & 99.09±0.71     \\
3 hidden layers &99.51±0.78 & 99.38±0.37 & 99.63±0.47 & 98.8±0.66     
\end{tabular}
\label{tbl:smaller_performance}
\end{table}
\begin{table}[t!]
\centering
\caption{The accuracies obtained by the neural networks in the tort law domain.}
\begin{tabular}{l|llll}
                & \multicolumn{4}{c}{\textbf{Trained on all instances}} \\
                & General    & Unique     & Unlawfulness & Imputability   \\ \hline
1 hidden layer  & 100±0      & 100±0      & 100±0      & 100±0      \\
2 hidden layers & 100±0      & 100±0      & 100±0      & 100±0     \\ 
3 hidden layers & 99.86±0.37 & 99.76±0.66 & 99.67±1.83 & 99.5±1.56 
\end{tabular}
\begin{tabular}{l|llll}
                & \multicolumn{4}{c}{\textbf{Trained on smaller dataset}} \\
                & General    & Unique     & Unlawfulness & Imputability   \\ \hline
1 hidden layer  & 98.45±0.5  & 97.24±0.89 & 92.8±3.47  & 91.22±4.04   \\
2 hidden layers & 99.03±0.44 & 98.27±0.78 & 95.71±3.1  & 94.38±3.84  \\ 
3 hidden layers & 98.23±0.72 & 96.83±1.28 & 92.96±5.33 & 91.45±3.51
\end{tabular}
\label{tbl:tort_performance}
\end{table}

\subsubsection{Accuracy}
Tables~\ref{tbl:new_performance} -- \ref{tbl:tort_performance} show the mean classification accuracies over 50 runs, together with their standard deviations, for the different combinations of training and testing sets in the three domains. These tables include the quantitatively measured performance on the various dedicated test sets.
Tables~\ref{tbl:new_performance} and~\ref{tbl:more_data_performance} present the results for the replication experiment with the original sizes of the Type A and Type B datasets and for the replication experiment with more data, respectively. The accuracies as reported in the orginal paper are provided in Table~\ref{tbl:original_performance} \citep{bench1993neural}). Results for the simplified welfare benefit domain are shown in Table~\ref{tbl:smaller_performance}, and for the tort law domain in Table~\ref{tbl:tort_performance}. 

\begin{figure*}[h!]
     \centering
     \begin{subfigure}[b]{0.3101\linewidth}
        \centering
        \includegraphics[width = 1 \linewidth]{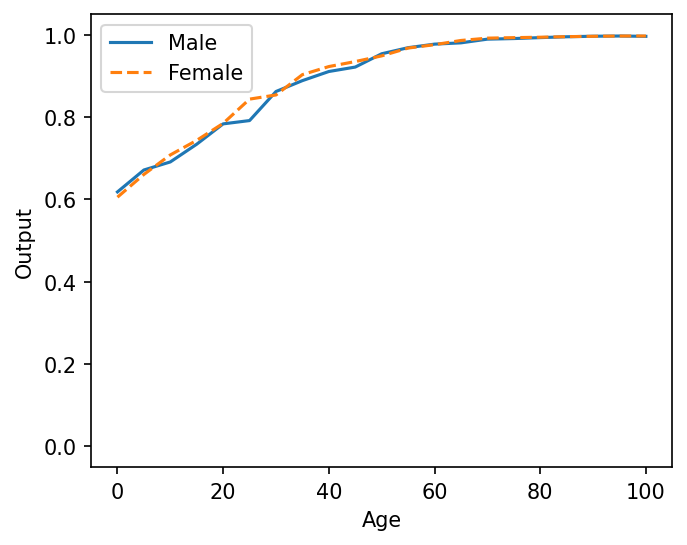}
        \caption{Replication}
        \label{fig:ag_A}
     \end{subfigure}
     ~
     \begin{subfigure}[b]{0.3101\linewidth}
        \centering
        \includegraphics[width = 1 \linewidth]{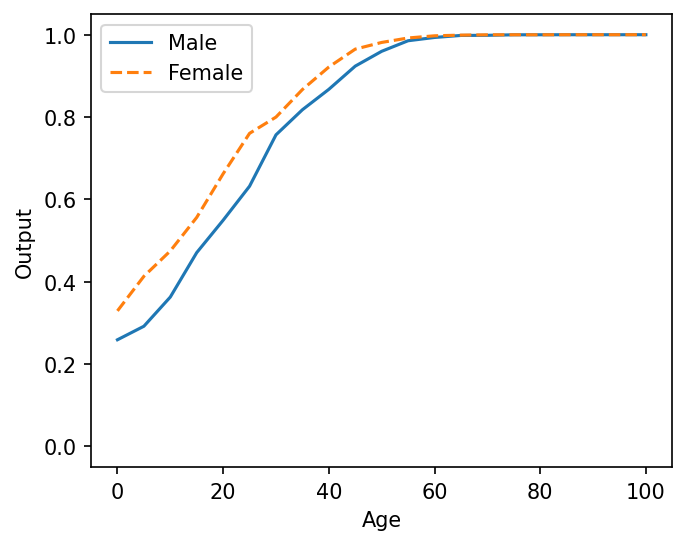}
        \caption{More data}
        \label{fig:ag_A_md}
     \end{subfigure}
     ~
     \begin{subfigure}[b]{0.3101\linewidth}
        \centering
        \includegraphics[width = 1 \linewidth]{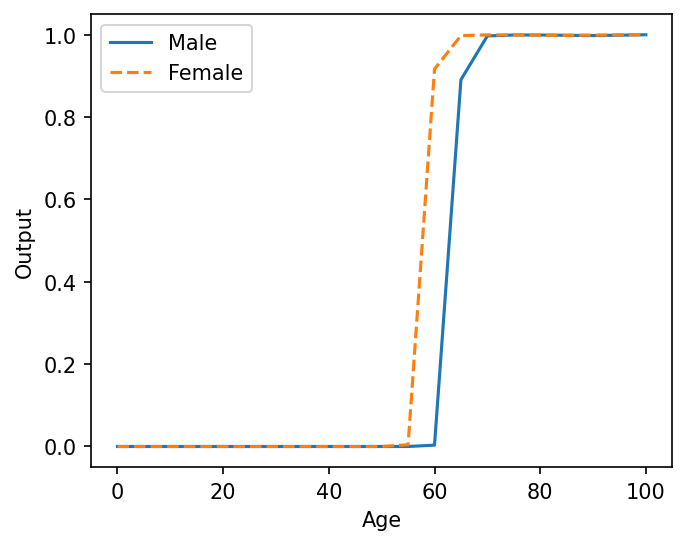}
        \caption{Simplified}
        \label{fig:simplified_ag_A}
     \end{subfigure}

     \begin{subfigure}[b]{0.3101\linewidth}
        \flushleft
        \includegraphics[width = 1 \linewidth]{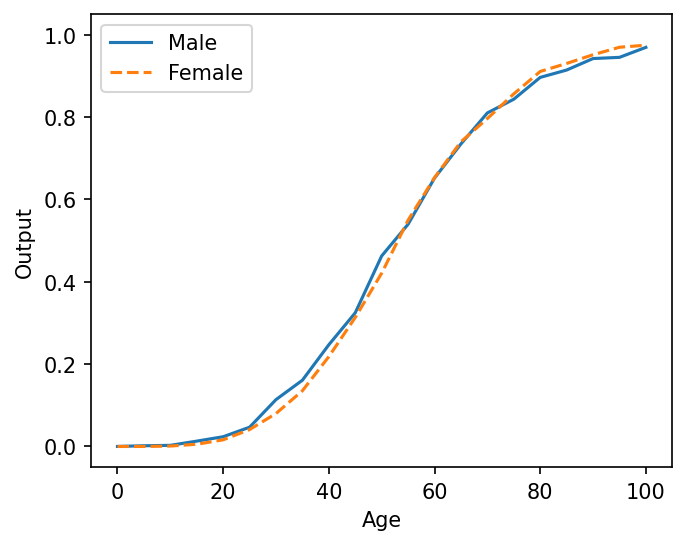}
        \caption{Replication}
        \label{fig:ag_B}
     \end{subfigure}
     ~
     \begin{subfigure}[b]{0.3101\linewidth}
        \flushleft
        \includegraphics[width = 1 \linewidth]{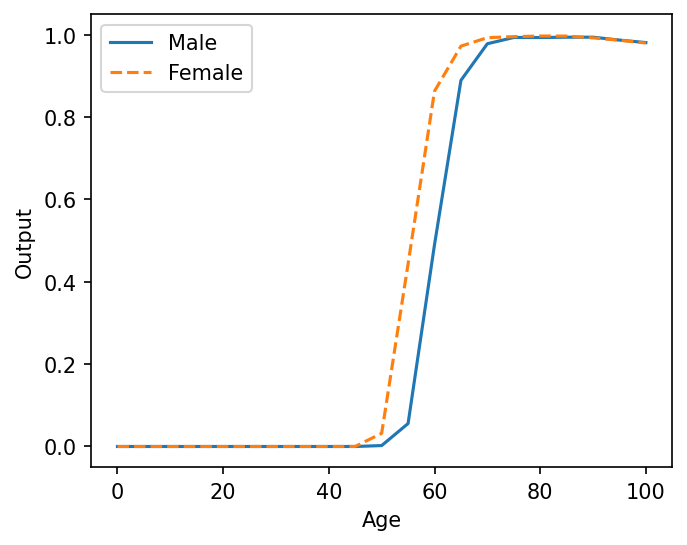}
        \caption{More data}
        \label{fig:ag_B_md}
     \end{subfigure}
     ~
     \begin{subfigure}[b]{0.3101\linewidth}
        \flushleft
        \includegraphics[width = 1 \linewidth]{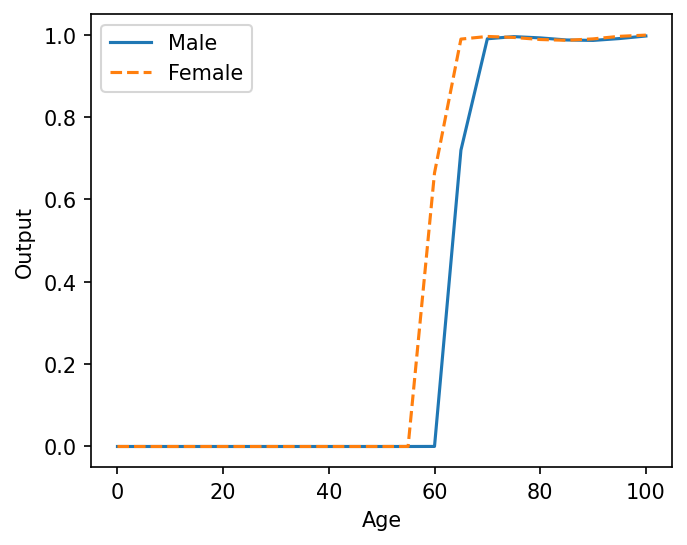}
        \caption{Simplified}
        \label{fig:simplified_ag_B}
     \end{subfigure}
    \caption{For all training sets from the (simplified) welfare domain: mean network output vs age on Age-Gender test set when trained on type A training sets (A-C) and on type B training sets (D-F).}
    \label{fig:ag}
\end{figure*}

\begin{figure*}[h!]
     \centering
     \begin{subfigure}[b]{0.3101\linewidth}
        \centering
        \includegraphics[width = 1 \linewidth]{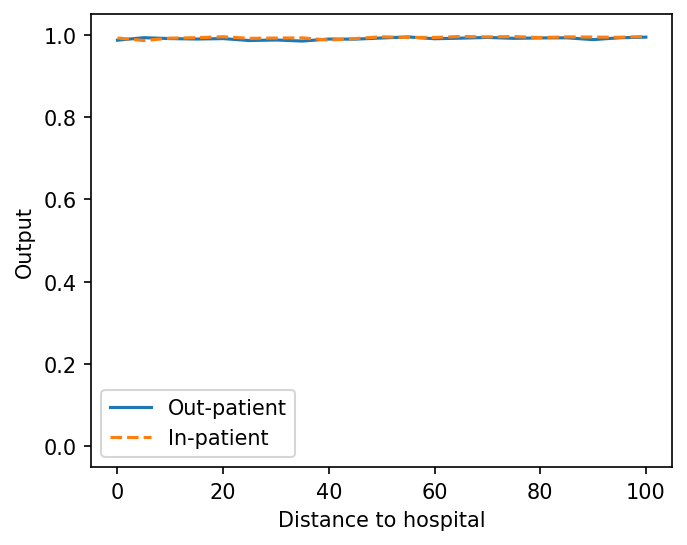}
        \caption{Replication}
        \label{fig:pd_A}
     \end{subfigure}
    ~
     \begin{subfigure}[b]{0.3101\linewidth}
        \centering
        \includegraphics[width = 1 \linewidth]{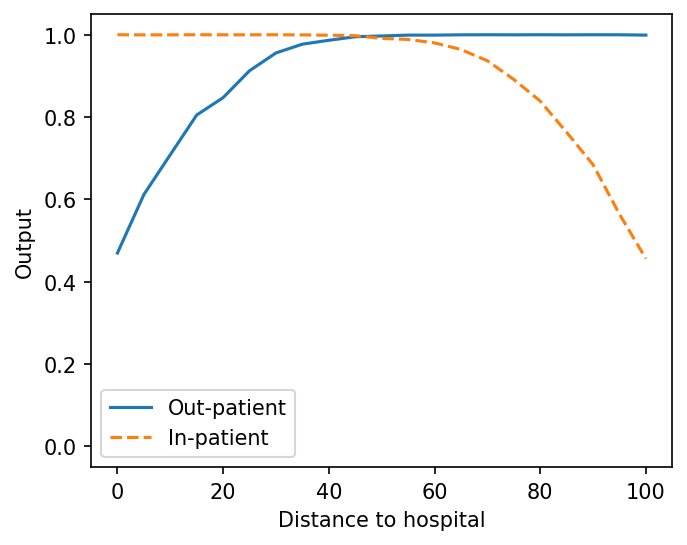}
        \caption{More data}
        \label{fig:pd_A_md}
     \end{subfigure}
    ~
     \begin{subfigure}[b]{0.3101\linewidth}
        \centering
        \includegraphics[width = 1 \linewidth]{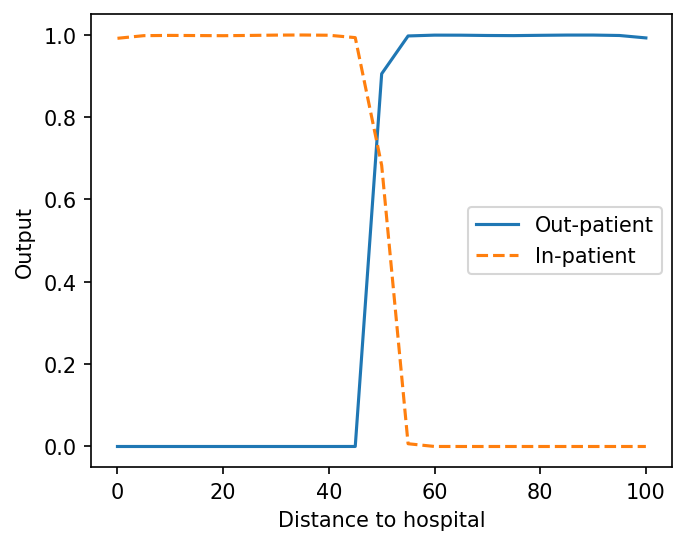}
        \caption{Simplified}
        \label{fig:simplified_pd_A}
     \end{subfigure}
     
     \begin{subfigure}[b]{0.3101\linewidth}
        \flushleft
        \includegraphics[width = 1 \linewidth]{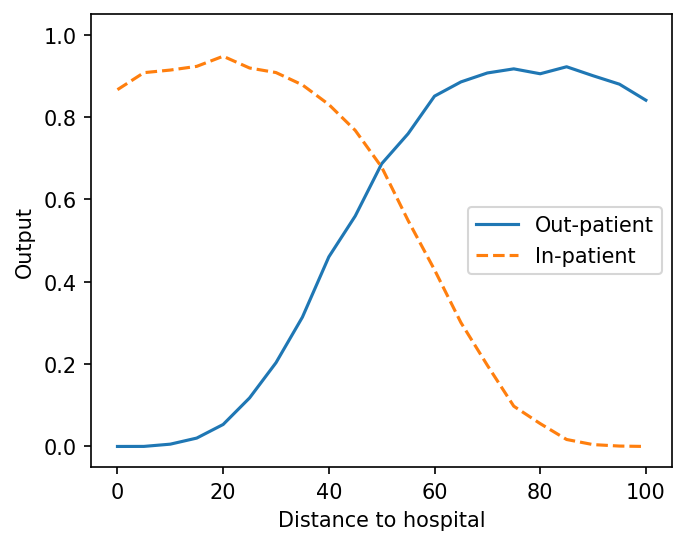}
        \caption{Replication}
        \label{fig:pd_B}
     \end{subfigure}
     ~
     \begin{subfigure}[b]{0.3101\linewidth}
        \flushleft
        \includegraphics[width = 1 \linewidth]{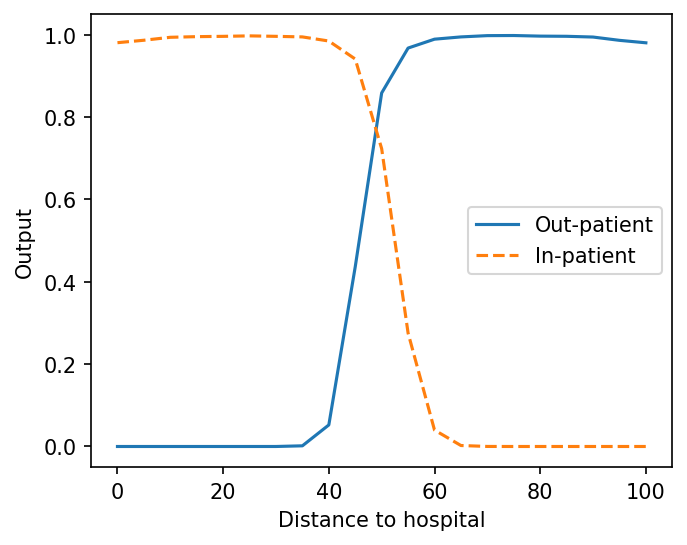}
        \caption{More data}
        \label{fig:pd_B_md}
     \end{subfigure}
     ~
     \begin{subfigure}[b]{0.3101\linewidth}
        \flushleft
        \includegraphics[width = 1 \linewidth]{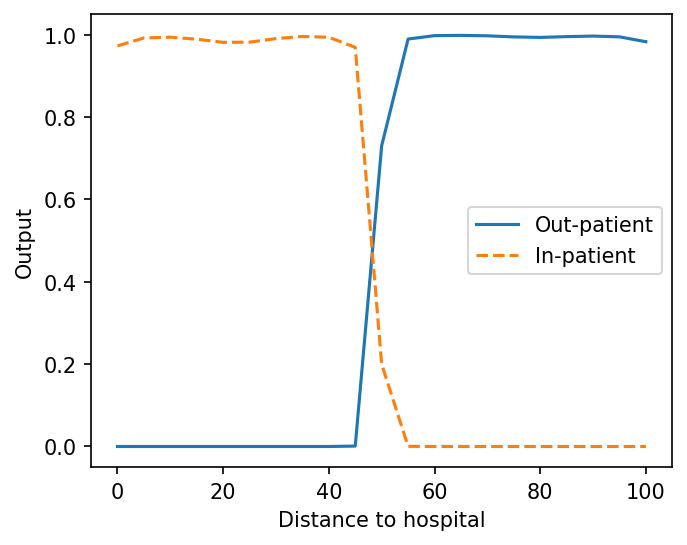}
        \caption{Simplified}
        \label{fig:simplified_pd_B}
     \end{subfigure}
    \caption{For all training sets from (simplified) welfare domain: mean network output vs distance on Patient-Distance test set when trained on type A training sets (A-C) and on type B training sets (D-F).}
    \label{fig:pd}
\end{figure*}

\subsubsection{Rationale discovery}
Each dedicated test dataset is designed to measure how well a model has learned a specific condition from the domain. Since performance on these test sets in terms of accuracy is comparable for the different neural network architectures used, we present the results for the qualitative evaluation of their rationale discovery capabilities for what is in theory the most sophisticated one: the models with 3 hidden layers.

In the welfare benefit domain, the Age-Gender datasets are used to measure how well condition $C_1$ is learned. In addition to measuring accuracy on these dedicated datasets, we can plot the actual output of the neural network, which should be 1 (eligible) for an individual of pensionable age and 0 otherwise, against age, for both values of gender. Such plots, showing the mean output of the networks over 50 runs, are shown in Figure~\ref{fig:ag} for both the welfare benefit domain and its simplified version, for networks trained on each of the training sets under consideration. 

Similarly, the Patient-Distance datasets are used to evaluate how well condition $C_6$ is learned, which can be assessed by plotting the networks' output against the distance to hospital, for both in-patients and out-patients. The plots showing the mean network output over 50 runs for both the welfare benefit domain and its simplified version, and for networks trained on each of the training sets under consideration, are shown in Figure~\ref{fig:pd}.

In the tort law domain, we can similarly evaluate how well conditions $c_2$ (unlawfullness) and $c_3$ (imputability) are learned. For these conditions, the network should output 1 in cases of the Unlawfulness dataset where the case is unlawful ($c_2$), or in the Imputability dataset where the case can be imputated to a person ($c_3$); otherwise the output should be 0. Since the tort law domain only contains Boolean features, the outputs of the networks are presented in tables rather than plots. The mean output over 50 runs for the two training sets on the Unlawfulness and Imputability datasets is presented in Table~\ref{tbl:tort_unlimp}. 

\section{Discussion}\label{sec:discussion}
In this section we discuss in detail the results we found and the conclusions we can draw from them. We separately focus on standard classification accuracy and on rational discovery capabilities. 
We conclude by introducing the approach we took as a general knowledge-driven method for model-agnostic rationale evaluation.

\subsection{Standard Accuracy}
Standard accuracy is measured to see whether the learned models are able to solve the classification problem, regardless of whether or not they discovered the rationale underlying the data. 

\subsubsection{Welfare benefit}
The accuracies obtained in the replication experiment (Table \ref{tbl:new_performance}) differ from those in the original study (Table \ref{tbl:original_performance}), but show similar trends. Originally, networks trained on a type A training set performed well on type A test sets (around 99\%), but much worse on the type B test set (around 70-76\%). When trained on a type B training set, the accuracies on test set A in the original study stayed the same, with accuracies on test set B increasing to around 98\%. In the replication experiment, training on a type B test set slightly decreases accuracies on type A test sets, while the accuracy on type B test sets increased less substantially than in the original experiment. In both cases, changing the distribution of the training data from type A to type B served to increase performance on test sets of the latter type while hardly affecting performance on test sets of the former type. Since type B datasets exploited some knowledge of the domain (i.e. benefit is typically denied due to failure on only a single condition), this suggests that overall performance can be improved using tailor made training sets. 

\begin{table}[t!]
\centering
\caption{Mean network output on the Unlawfulness and Imputability datasets versus the logical evaluation of the unlawfulness resp. imputability conditions.}
\begin{tabular}{ll|ll}
\multicolumn{2}{l|}{\textbf{Trained on all instances}} & \multicolumn{2}{l}{\textbf{Trained on smaller dataset}} \\ \hline
\textbf{Unlawfulness}    & \textbf{Output}   & \textbf{Unlawfulness}     & \textbf{Output}              \\
False                    & 0                 & False                     & 0.018               \\
True                     & 1                 & True                      & 1\\
\hline
\textbf{Imputability}    & \textbf{Output}   & \textbf{Imputability}     & \textbf{Output}              \\
False                    & 0                 & False                     & 0.875               \\
True                     & 1                 & True                      & 1  
\end{tabular}
\label{tbl:tort_unlimp}
\end{table}

Using more data, we find higher accuracies (Table \ref{tbl:more_data_performance}). This is not surprising, as more training data generally leads to a better performance. Still, for networks trained on type A data sets, accuracies on a type B test sets (below 85\%) are much lower than on type A test sets (around 99.8\%). Training the networks on type B training sets with more data shows significantly better results than with fewer datapoints. Accuracies on type A test sets then are still around 99\%, whereas the accuracies on type B test sets are around 98\%. The above observations therefore still hold, even with more data and modern machine learning methods. 

\subsubsection{Simplified}
In the simplified domain, high accuracies are found across all datasets, averaging out at around 99\% (Table \ref{tbl:smaller_performance}). Accuracies on type B test sets are only slightly lower than on type A test sets, unlike in the other welfare domain experiments. The networks do seem to perform slightly better on both types of test sets when trained on a type B training set as compared to a type A training set. However, this difference is much more nuanced than in the regular welfare benefit dataset. This can be explained by the fact that cases in type A data sets can now fail on at most 2 conditions, rather than the original 6, which is only 1 more than the single failed condition in the type B datasets.

\subsubsection{Tort law}
In the Tort law domain we find accuracies of 100\% or near 100\% for networks trained on all instances (see Table \ref{tbl:tort_performance}). When presented with all unique instances, the networks with one and two hidden layers are able to perfectly predict the outcome from the Dutch tort law, and the network with three hidden layers can create a very close approximation. 

Presenting a neural network with all available cases is in practice often infeasible. If it is possible, then a simple lookup table rather than a neural network would most likely suffice. For this reason, we also trained the networks on a subset of only around 35\% of the unique instances (see Table \ref{tbl:tort_performance}). As expected, the accuracies of the networks on the general test sets drop, but only slightly (to 98-99\%). Even on the unique test set, accuracies remain around 96\%. This suggests that it is possible to approximate tort law with a small subset of the unique cases. 

\subsection{Rationale Discovery}
Looking at the performance of the networks on the dedicated test sets partially exposes the rationale captured by the network. We designed these test sets such that each one targets a single condition from the domain. 
In addition to considering the accuracy on these dedicated test sets, we qualitatively evaluate the rational discovery capabilities of the networks by comparing their outputs with the actual outputs we would ideally expect for the different domains.

\subsubsection{Welfare benefit}
\begin{figure}[b!]
    \centering
    \includegraphics[width = 1 \linewidth]{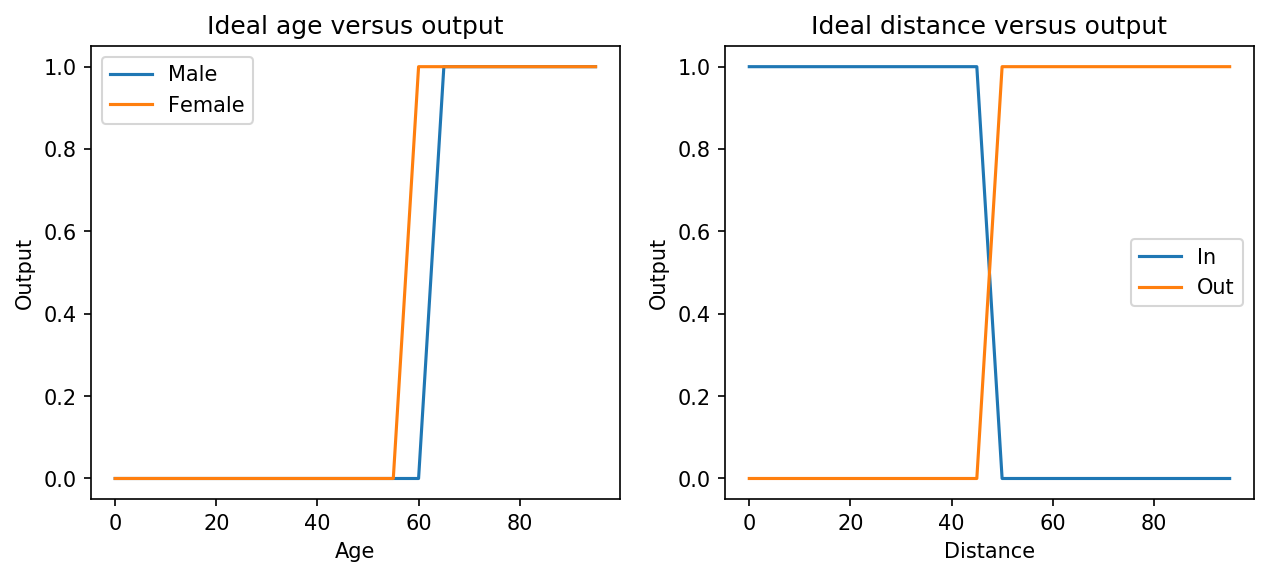}
    \caption{An idealistic expectation of the outputs of a network on the Age-Gender dataset versus the age for both genders (left) and on the Patient-Distance dataset versus the distance for both patient types (right).}
    \label{fig:ideal_age_distance}
\end{figure}

In the welfare benefit domain, the Age-Gender dataset is used to measure how well condition $C_1$ is learned, that is, whether the networks output 1 if the individual is of pensionable age (male and over the age of 65 or female and over the age of 60), and output 0 otherwise. Plotting the age of the individuals from the Age-Gender dataset against the output of the network, for each gender, should ideally result in the graph on the left side of Figure~\ref{fig:ideal_age_distance}. Here the output of the network spikes instantly from 0 to 1 at the age of 60 for women and 65 for men.

Similarly, for cases from the Patient-Distance dataset the networks should only output 1 (eligible) if the relative is an in-patient and the distance to the hospital is less than 50 miles, or if the relative is an out-patient and the distance to the hospital is further than 50 miles (condition $C_6$). Plotting the distance against the output of the network for both types of patients would ideally result in the graph shown on the right in Figure~\ref{fig:ideal_age_distance}.

In our replication experiment the output graphs show a similar pattern as in the original 1993 experiment. In the latter (not shown), the networks trained on a type A training set do not show the expected pattern, for neither condition; in fact for the Patient-Distance test cases always a 1 is returned. Training on a type B training set improved the results, but the turning point at which the networks output 1 is off. For the Age-Gender dataset it occurs at 45 for women, rather than 60, and at 50 for men, instead of 65. For the Patient-Distance dataset the turning point was too gradual, and takes place at 40, rather than 50 miles. 
In our replication experiment the outputs of the networks trained on a type B dataset (Figures~\ref{fig:ag}(D) and \ref{fig:pd}(D)) more closely resemble the ideal outputs than the outputs of the networks trained on a type A dataset (Figures~\ref{fig:ag}(A) and \ref{fig:pd}(A)), for both conditions. For the patient-distance condition ($C_6$), the turning point does occur at 50 after training on a type B dataset, unlike in the original experiment. 
Training on a type B dataset seems to have a significant impact on the way the rationale of the networks is formed, as networks are able to internalize condition $C_1$ and $C_6$ better when trained on a type B dataset. This is furthermore reflected in the accuracies on the Age-Gender and patient distance dataset as shown in Table \ref{tbl:new_performance}, which increase by roughly 30\% when training on a type B dataset. These accuracies on the Age-Gender and Patient-Distance datasets were not present in the original study.

Upon repeating the replication experiment with more data, this indeed increases the performance of the networks significantly, but we still find performance for networks trained on type B datasets to be better than that of networks trained on  type A datasets. Output patterns more closely resemble the ideal ones after training on type B datasets (see Figure~\ref{fig:ag}(B) versus (E) and Figure~\ref{fig:pd}(B) versus (E)) and accuracies also increase (see Table \ref{tbl:more_data_performance}). Interestingly, the turning point for condition $C_1$ does occur at the right place when training on more type B training data: at 60 for females and 65 for males (see Figure~\ref{fig:ag}(E)).

\subsubsection{Simplified}
The simplified domain consists of only the two conditions $C_1$ and $C_6$, without any other conditions or noise variables. In this less complex version of the domain, overall performance is much higher, and conditions $C_1$ and $C_6$ are learned quite successfully. Figures~\ref{fig:ag}(C) and (F), and \ref{fig:pd}(C) and (F), respectively, are very close to the ideal output graphs, with turning points in the correct places. This is also reflected in near perfect accuracy scores on Age-Gender and Patient-Distance datasets in Table \ref{tbl:smaller_performance}. 
As argued before, the difference between type A and type B datasets is much smaller than in the original domain, hence the results found for these to datasets are now quite similar. 

\subsubsection{Tort law}
Recall that in the tort law domain, on the Imputability dataset, networks should output 1 if the case can be imputated to the person, and 0 otherwise; on the Unlawfulness dataset, the networks should output 1 if the case is unlawful, and 0 otherwise. Table~\ref{tbl:tort_unlimp} shows well the networks were able to internalize the notions of unlawfulness and imputability. When trained on all instances, the mean output of the networks is 0 if a case is not unlawful, and 1 if it is, which is exactly what it should do. Networks trained on all instances attain a perfect score on the Imputability dataset as well. This can also be seen in Table \ref{tbl:tort_performance}, where the networks score 100\% accuracy on the Unlawfulness and Imputability datasets after training on all instances.

With less data, however, accuracies drop to around 92-95\% for the Unlawfulness dataset and 91-94\% for the Imputability dataset. This accuracy may still seem high, but we should take into account the label distributions (66.67-33.33\% and 87.5-12.5\%, respectively). Table~\ref{tbl:tort_unlimp} shows that networks still perform perfectly on cases in which the unlawfulness and imputability conditions evaluate to true. When the conditions are false, however, mistakes are made. The average output of networks on the Unlawfulness dataset increases to 0.018, which should be 0, meaning that it classifies some lawful cases as unlawful. In the Imputability dataset, the mean output increased more drastically to 0.875 when imputability is false. Meaning that in 87.5\% of the instances in which the case cannot be imputed to a person, the network incorrectly decided that it should. This means that despite its high accuracy on the general test set, the networks largely ignored the concept of imputability. 

\subsection{A Method for Rationale Evaluation}
Although our experiments and discussion focused on specific example domains and neural networks, our approach for rationale evaluation can be seen as a general method independent of the machine learning algorithm applied. This paper therefore proposes a 
knowledge-driven method for model-agnostic rationale evaluation, consisting of three distinct steps:

\begin{enumerate}
    \item Measure the accuracy of a trained system, and proceed if the accuracy is sufficiently high;
    \item Design dedicated test sets for rationale evaluation targeting selected rationale elements based on expert knowledge of the domain;
    \item Evaluate the rationale through the performance of the trained system on these dedicated test sets.
\end{enumerate}
\noindent
The first step is based on the assumption that efforts for assessing and possibly improving the rationale discovery capabilities of a learned model are only taken if the general performance of the model is already considered good enough. Here we assume performance is measured using accuracy, but other measures can be employed as well and the threshold of what is considered good enough may vary per domain and application. 

The second step in our method depends on domain knowledge. Hence the method effectively is a quantitative human-in-the-loop solution for rationale evaluation.

In the third step, performance is again evaluated, by now not only considering accuracy but also examining model output and expected output in terms of the dedicated test sets. Our examples have shown that the latter depend on the type of features involved. 

Subsequently, the information gained by using this rationale evaluation method can be used to improve the rationale of the system by adjusting the training data accordingly, imposing sound rationale discovery. 

The method does not currently specify how the dedicated test sets are constructed. We aim to further operationalize the rationale evaluation method by using information about the knowledge in the domain, and the distribution of examples, for instance building on Bayesian networks.

\section{Conclusion}
The work in this paper was inspired by Bench-Capon's 1993 paper that investigated whether neural networks are able to tackle open texture problems. The conclusions were that neural networks can perform very well on such problems in terms of accuracy, even if some conditions from the domain are not learned~\citep{bench1993neural}. 

In this paper we first replicated the original experiments as closely as possible to verify that we can reproduce the results from the 1993 paper. In addition, we repeated the experiments with larger training datasets to ensure that the original conclusions about conditions that were not learned are not due to a lack of data. The idea of constructing test cases to test specific conditions inspired us to propose a method for assessing rationale discovery capabilities by designing dedicated test datasets and to evaluate performance on these knowledge-driven test sets, combining quantitative and qualitative evaluation elements in a hybrid way. Type B datasets served to complicate the problem in the original study, but also demonstrate that training can be improved using knowledge-driven tailor made training sets. 

We investigated three legal domains, in which neural networks were trained on labelled cases and tasked with predicting unlabelled cases. We started off with an artificial domain from the literature, followed by a simplified adaptation of that domain. Lastly, we investigated a real life domain as well. %
The results indicate that the network are able to achieve high accuracies in each of the three domains. The networks are therefore able to make the right decisions in most cases, with accuracies averaging around 99\% on type A or regular test sets. This is how machine learning problems are usually evaluated. 
Using our approach of rationale evaluation, however, we show that the networks do not necessarily learn the conditions, despite their high accuracy scores. Performance on the dedicated test sets, type B, Age-Gender and Patient-Distance dataset show that the networks are unable to learn the conditions $C_1$ and $C_6$. This was suggested in the original experiment \citep{bench1993neural} and it holds true in the replication study with modern, commonly used machine learning techniques and more data. By adjusting the distribution of the training data based on expert domain knowledge (training on a type B dataset) these accuracies increase. Simplifying the domain shows that systems are able to learn the conditions $C_1$ and $C_6$, though still not perfectly. Even in the real life tort law domain, with a non-fictional knowledge structure and different characteristics, a similar pattern can be observed. The networks failed to learn the independent condition that defines imputability, despite its high accuracies on the general test set. 

This study therefore reaffirms the conclusions from previous work, while simultaneously introducing a model-agnostic method for assessing rationale discovery capabilities of machine learned black box models, using dedicated test datasets designed with expert knowledge of the domain. In future research, we aim to further detail and extend our method such that by employing it, the soundness of the rationale becomes tangible, and its quality can be asserted. Ultimately, based on this evaluation, the training data of the black-box systems can be altered to improve their rationale. Further expanding upon this design method will bring us closer to AI that is both explainable and responsible. 

%
%
%

\bibliographystyle{splncs04}
\bibliography{sample-base}

\begin{thebibliography}{30}
\providecommand{\natexlab}[1]{#1}
\providecommand{\url}[1]{\texttt{#1}}
\expandafter\ifx\csname urlstyle\endcsname\relax
  \providecommand{\doi}[1]{doi: #1}\else
  \providecommand{\doi}{doi: \begingroup \urlstyle{rm}\Url}\fi

\bibitem[Ashley(1990)]{ashley1990}
K.~D. Ashley.
\newblock \emph{Modeling Legal Arguments: Reasoning with Cases and
  Hypotheticals}.
\newblock The MIT Press, Cambridge (Massachusetts), 1990.

\bibitem[Ashley(2019)]{ashley2019}
K.~D. Ashley.
\newblock A brief history of the changing roles of case prediction in ai and
  law.
\newblock \emph{Law in Context}, 36\penalty0 (1):\penalty0 93--112, 2019.

\bibitem[Atkinson et~al.(2020{\natexlab{a}})Atkinson, Bench-Capon, Bex, Gordon,
  Prakken, Sartor, and Verheij]{atkinson2020memoriam}
K.~Atkinson, T.~Bench-Capon, F.~Bex, T.~F. Gordon, H.~Prakken, G.~Sartor, and
  B.~Verheij.
\newblock In memoriam douglas n. walton: the influence of doug walton on ai and
  law.
\newblock \emph{Artificial Intelligence and Law}, pages 1--46,
  2020{\natexlab{a}}.

\bibitem[Atkinson et~al.(2020{\natexlab{b}})Atkinson, Bench-Capon, and
  Bollegala]{atkinson2020explanation}
K.~Atkinson, T.~Bench-Capon, and D.~Bollegala.
\newblock Explanation in ai and law: Past, present and future.
\newblock \emph{Artificial Intelligence}, 289:\penalty0 103387,
  2020{\natexlab{b}}.

\bibitem[Bench-Capon(1993)]{bench1993neural}
T.~Bench-Capon.
\newblock Neural networks and open texture.
\newblock In \emph{{Proceedings of the 4th International Conference on
  Artificial Intelligence and Law}}, ICAIL '93, pages 292--297. ACM, New York,
  1993.
\newblock ISBN 0-89791-606-9.

\bibitem[Br\"{u}ninghaus and Ashley(2003)]{bruninghausAshley2003}
S.~Br\"{u}ninghaus and K.~D. Ashley.
\newblock Predicting outcomes of case based legal arguments.
\newblock In \emph{Proceedings of the 9th {I}nternational {C}onference on
  {A}rtificial {I}ntelligence and {L}aw (ICAIL 2003)}, pages 233--242. ACM, New
  York (New York), 2003.

\bibitem[Goodfellow et~al.(2015)Goodfellow, Shlens, and
  Szegedy]{goodfellow2014explaining}
I.~J. Goodfellow, J.~Shlens, and C.~Szegedy.
\newblock Explaining and harnessing adversarial examples.
\newblock In \emph{Proceedings of International Conference on Learning
  Representations}, 2015.

\bibitem[Gordon(1995)]{gordon1995}
T.~F. Gordon.
\newblock \emph{The Pleadings Game: An Artificial Intelligence Model of
  Procedural Justice}.
\newblock Kluwer, Dordrecht, 1995.

\bibitem[Grabmair et~al.(2015)Grabmair, Ashley, Chen, Sureshkumar, Wang,
  Nyberg, and Walker]{grabmairEtal2015}
M.~Grabmair, K.~D. Ashley, R.~Chen, P.~Sureshkumar, C.~Wang, E.~Nyberg, and
  V.~R. Walker.
\newblock Introducing {LUIMA}: an experiment in legal conceptual retrieval of
  vaccine injury decisions using a uima type system and tools.
\newblock In \emph{Proceedings of the 15th International Conference on
  Artificial Intelligence and Law}, pages 69--78. ACM, New York (New York),
  2015.

\bibitem[Hage et~al.(1993)Hage, Leenes, and Lodder]{hageLeenesLodder1993}
J.~C. Hage, R.~Leenes, and A.~R. Lodder.
\newblock Hard cases: a procedural approach.
\newblock \emph{Artificial intelligence and law}, 2\penalty0 (2):\penalty0
  113--167, 1993.

\bibitem[Hunter(1999)]{hunter1999out}
D.~Hunter.
\newblock Out of their minds: Legal theory in neural networks.
\newblock \emph{Artificial Intelligence and Law}, 7\penalty0 (2):\penalty0
  129--151, 1999.

\bibitem[Kingma and Ba(2015)]{kingma2014adam}
D.~P. Kingma and J.~Ba.
\newblock Adam: {A} method for stochastic optimization.
\newblock In \emph{Proceedings of 3rd International Conference on Learning
  Representations}, 2015.

\bibitem[Leighton and Wieland(1994)]{Leighton1994}
R.~R. Leighton and A.~P. Wieland.
\newblock \emph{The Aspirin/Migraines Software Package}, pages 209--227.
\newblock Springer, New York, Boston, MA, 1994.
\newblock ISBN 978-1-4615-2736-7.

\bibitem[Lundberg and Lee(2017)]{NIPS2017_7062}
S.~M. Lundberg and S.~Lee.
\newblock A unified approach to interpreting model predictions.
\newblock In \emph{Advances in Neural Information Processing Systems 30}, pages
  4765--4774. Curran Associates, Inc., 2017.

\bibitem[Medvedeva et~al.(2019)Medvedeva, Vols, and Wieling]{medvedevaEtal2019}
M.~Medvedeva, M.~Vols, and M.~Wieling.
\newblock Using machine learning to predict decisions of the european court of
  human rights.
\newblock \emph{Artificial Intelligence and Law}, pages 1--30, 2019.

\bibitem[Miller(2019)]{MILLER20191}
T.~Miller.
\newblock Explanation in artificial intelligence: Insights from the social
  sciences.
\newblock \emph{Artificial Intelligence}, 267:\penalty0 1--38, 2019.
\newblock ISSN 0004-3702.

\bibitem[Mochales~Palau and Moens(2009)]{mochalesMoens2009}
R.~Mochales~Palau and M.~F. Moens.
\newblock Argumentation mining: the detection, classification and structure of
  arguments in text.
\newblock In \emph{Proceedings of the 12th {I}nternational {C}onference on
  {A}rtificial {I}ntelligence and {L}aw (ICAIL 2009)}, pages 98--107. ACM
  Press, New York (New York), 2009.

\bibitem[Mo{\v{z}}ina et~al.(2005)Mo{\v{z}}ina, {\v{Z}}abkar, Bench-Capon, and
  Bratko]{movzina2005argument}
M.~Mo{\v{z}}ina, J.~{\v{Z}}abkar, T.~Bench-Capon, and I.~Bratko.
\newblock Argument based machine learning applied to law.
\newblock \emph{Artificial Intelligence and Law}, 13\penalty0 (1):\penalty0
  53--73, 2005.

\bibitem[Pedregosa et~al.(2011)Pedregosa, Varoquaux, Gramfort, Michel, Thirion,
  Grisel, Blondel, Prettenhofer, Weiss, Dubourg, Vanderplas, Passos,
  Cournapeau, Brucher, Perrot, and Duchesnay]{scikit-learn}
F.~Pedregosa, G.~Varoquaux, A.~Gramfort, V.~Michel, B.~Thirion, O.~Grisel,
  M.~Blondel, P.~Prettenhofer, R.~Weiss, V.~Dubourg, J.~Vanderplas, A.~Passos,
  D.~Cournapeau, M.~Brucher, M.~Perrot, and E.~Duchesnay.
\newblock Scikit-learn: machine learning in {P}ython.
\newblock \emph{Journal of Machine Learning Research}, 12:\penalty0 2825--2830,
  2011.

\bibitem[Philipps and Sartor(1999)]{philipps1999introduction}
L.~Philipps and G.~Sartor.
\newblock Introduction: from legal theories to neural networks and fuzzy
  reasoning.
\newblock \emph{Artificial Intelligence and law}, 7\penalty0 (2):\penalty0
  115--128, 1999.

\bibitem[Ribeiro et~al.(2016)Ribeiro, Singh, and Guestrin]{lime}
M.~T. Ribeiro, S.~Singh, and C.~Guestrin.
\newblock "why should {I} trust you?": Explaining the predictions of any
  classifier.
\newblock In \emph{Proceedings of the 22nd {ACM} {SIGKDD} International
  Conference on Knowledge Discovery and Data Mining, San Francisco, CA, USA},
  pages 1135--1144, 2016.

\bibitem[Rissland and Ashley(1987)]{HYPO}
E.~L. Rissland and K.~D. Ashley.
\newblock A case-based system for trade secrets law.
\newblock In \emph{Proceedings of the 1st International Conference on
  Artificial Intelligence and Law}, ICAIL '87, pages 60--66, New York, NY, USA,
  1987. ACM.
\newblock ISBN 0897912306.

\bibitem[Steging et~al.(2021)Steging, Renooij, and Verheij]{ICAILpaper}
C.~Steging, S.~Renooij, and B.~Verheij.
\newblock Discovering the rationale of decisions: Towards a method for aligning
  learning and reasoning (accepted).
\newblock In \emph{Proceedings of the 18th International Conference on
  Artificial Intelligence and Law}, ICAIL '21. ACM, New York, 2021.

\bibitem[Stranieri et~al.(1999)Stranieri, Zeleznikow, Gawler, and
  Lewis]{stranieriEtal1999}
A.~Stranieri, J.~Zeleznikow, M.~Gawler, and B.~Lewis.
\newblock A hybrid rule--neural approach for the automation of legal reasoning
  in the discretionary domain of family law in australia.
\newblock \emph{Artificial Intelligence and Law}, 7\penalty0 (2-3):\penalty0
  153--183, 1999.

\bibitem[Verheij(2003{\natexlab{a}})]{verheij2003aaa}
B.~Verheij.
\newblock Artificial argument assistants for defeasible argumentation.
\newblock \emph{Artificial Intelligence}, 150\penalty0 (1--2):\penalty0
  291--324, 2003{\natexlab{a}}.

\bibitem[Verheij(2003{\natexlab{b}})]{verheij2003dialectical}
B.~Verheij.
\newblock Dialectical argumentation with argumentation schemes: An approach to
  legal logic.
\newblock \emph{Artificial intelligence and Law}, 11\penalty0 (2-3):\penalty0
  167--195, 2003{\natexlab{b}}.

\bibitem[Verheij(2017)]{verheij2017formalizing}
B.~Verheij.
\newblock Formalizing arguments, rules and cases.
\newblock In \emph{Proceedings of the 16th International Conference on
  Artificial Intelligence and Law}, ICAIL '17, pages 199--208. ACM, New York,
  2017.
\newblock ISBN 978-1-4503-4891-1.

\bibitem[Vlek et~al.(2016)Vlek, Prakken, Renooij, and Verheij]{vlek2016method}
C.~S. Vlek, H.~Prakken, S.~Renooij, and B.~Verheij.
\newblock A method for explaining bayesian networks for legal evidence with
  scenarios.
\newblock \emph{Artificial Intelligence and Law}, 24\penalty0 (3):\penalty0
  285--324, 2016.

\bibitem[Wardeh et~al.(2009)Wardeh, Bench-Capon, and Coenen]{wardeh2009padua}
M.~Wardeh, T.~Bench-Capon, and F.~Coenen.
\newblock Padua: a protocol for argumentation dialogue using association rules.
\newblock \emph{Artificial Intelligence and Law}, 17\penalty0 (3):\penalty0
  183--215, 2009.

\bibitem[Wyner et~al.(2010)Wyner, Mochales-Palau, Moens, and
  Milward]{wynerEtal2010}
A.~Wyner, R.~Mochales-Palau, M.~F. Moens, and D.~Milward.
\newblock Approaches to text mining arguments from legal cases.
\newblock In \emph{Semantic Processing of Legal Texts}, pages 60--79. Springer,
  Berlin, 2010.

\end{thebibliography}

\end{document}